\documentclass[11pt,a4paper]{article}
\usepackage[hyperref]{emnlp-ijcnlp-2019}
\usepackage{times}
\usepackage{tabularx}
\usepackage{latexsym}
\usepackage{graphicx}
\usepackage{caption}
\usepackage{subcaption}
\usepackage{float}
\usepackage{booktabs}
\usepackage{listings}
\usepackage{url}
\usepackage{color}
\usepackage{enumitem}
\usepackage{tikz}
\usepackage{lipsum}
\newcommand\blfootnote[1]{%
  \begingroup
  \renewcommand\thefootnote{}\footnote{#1}%
  \addtocounter{footnote}{-1}%
  \endgroup
}

\def\checkmark{\tikz\fill[scale=0.4](0,.35) -- (.25,0) -- (1,.7) -- (.25,.15) -- cycle;} 
\definecolor{codegreen}{rgb}{0,0.6,0}
\definecolor{codegray}{rgb}{0.5,0.5,0.5}
\definecolor{codepurple}{rgb}{0.58,0,0.82}
\definecolor{backcolour}{rgb}{0.95,0.95,0.92}
 
\lstdefinestyle{pythonstyle}{
    backgroundcolor=\color{backcolour},   
    commentstyle=\color{codegreen},
    keywordstyle=\color{magenta},
    numberstyle=\tiny\color{codegray},
    stringstyle=\color{codepurple},
    basicstyle=\ttfamily,
    breakatwhitespace=false,         
    breaklines=true,                 
    captionpos=b,                    
    keepspaces=true,                 
    numbers=left,                    
    numbersep=5pt,                  
    showspaces=false,                
    showstringspaces=false,
    showtabs=false,                  
    tabsize=2,
}
\aclfinalcopy 

\title{VizSeq: A Visual Analysis Toolkit for Text Generation Tasks}

\author{Changhan Wang$^{\dagger}$, Anirudh Jain*$^{\ddagger}$, Danlu Chen$^{\dagger}$ and Jiatao Gu$^{\dagger}$ \\
  $^{\dagger}$ Facebook AI Research, $^{\ddagger}$ Stanford University \\
  {\tt \{changhan, danluchen, jgu\}@fb.com, anirudhj@stanford.edu}}
\date{}

\begin{document}
\maketitle
\blfootnote{\textbf{*} Work carried out during an internship at Facebook.}
\begin{abstract}
Automatic evaluation of text generation tasks (e.g. machine translation, text summarization, image captioning and video description) usually relies heavily on 
task-specific metrics, such as BLEU~\citep{papineni2002bleu} and ROUGE~\citep{lin2004rouge}. They, however, are abstract numbers and are not perfectly aligned with human assessment. This suggests inspecting detailed examples as a complement to identify system error patterns. In this paper, we present VizSeq, a visual analysis toolkit for instance-level and corpus-level system evaluation on a wide variety of text generation tasks. It supports multimodal sources and multiple text references, providing visualization in Jupyter notebook or a web app interface. It can be used locally or deployed onto public servers for centralized data hosting and benchmarking. It covers most common n-gram based metrics accelerated with multiprocessing, and also provides latest embedding-based metrics such as BERTScore~\cite{zhang2019bertscore}.
\end{abstract}

\section{Introduction}
Many natural language processing (NLP) tasks can be viewed as conditional text generation problems, where natural language texts are generated given inputs in the form of text (e.g. machine translation), image (e.g. image captioning), audio (e.g. automatic speech recognition) or video (e.g. video description). Their automatic evaluation usually relies heavily on task-specific metrics. 
Due to the complexity of natural language expressions, those metrics are not always perfectly aligned with human assessment. Moreover, metrics only produce abstract numbers and are limited in illustrating system error patterns. This suggests the necessity of inspecting detailed evaluation examples to get a full picture of system behaviors as well as seek improvement directions.

\begin{figure}[t]
    \centering
    \small
    \includegraphics[width=0.48\textwidth]{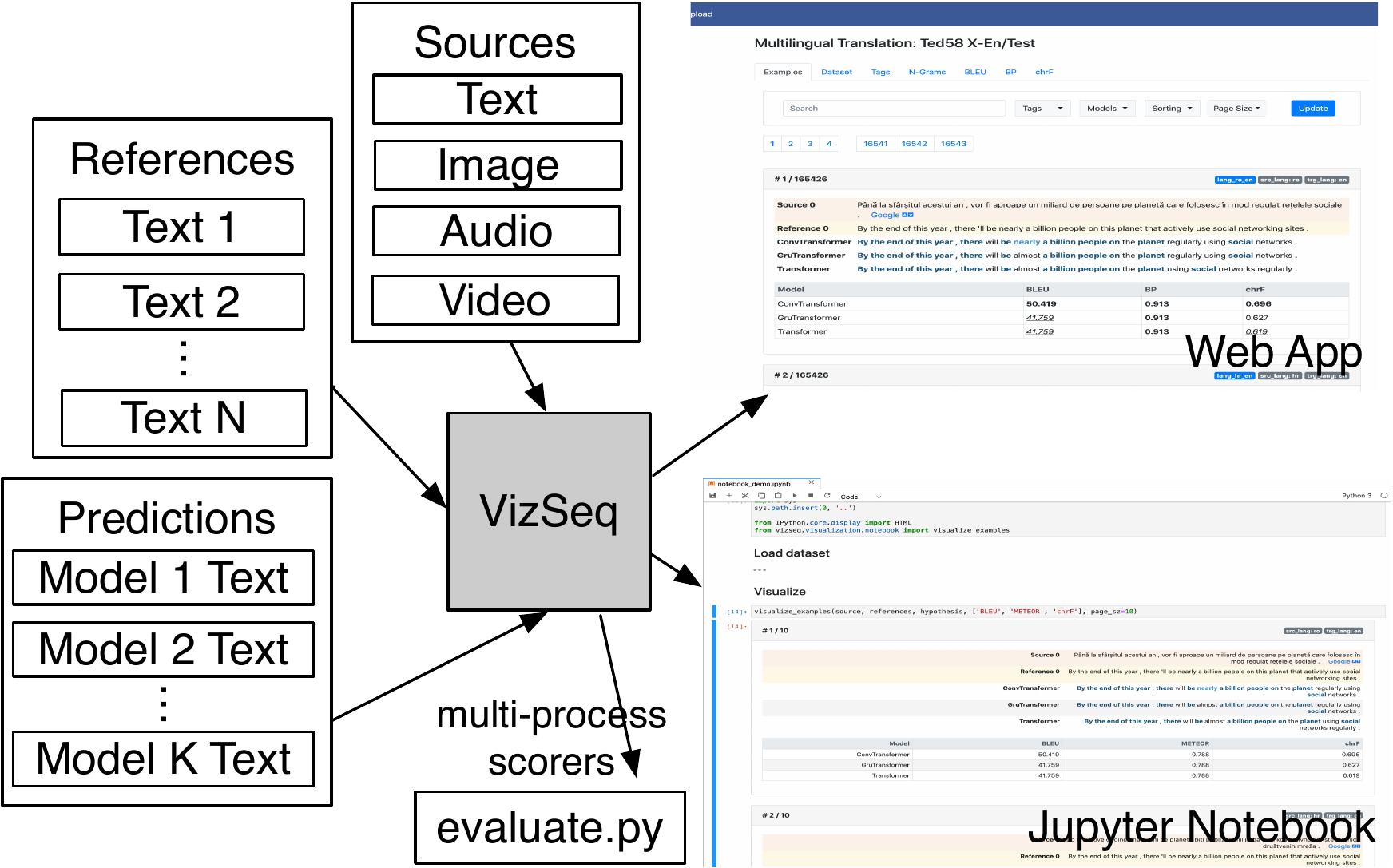}
    \caption{An overview of VizSeq. VizSeq takes multimodal sources, text references as well as model predictions as inputs, and analyzes them visually in Jupyter notebook or in a web app interface. It can also be used without visualization as a normal Python package.}
    \label{fig:vizseq_overview}
    \vspace{-12pt}
\end{figure}

A bunch of softwares have emerged to facilitate calculation of various metrics or demonstrating examples with sentence-level scores in an integrated user interface: ibleu~\citep{madnani2011ibleu}, MTEval~\footnote{https://github.com/odashi/mteval}, MT-ComparEval~\citep{klejch2015mt}, nlg-eval~\citep{sharma2017nlgeval}, Vis-Eval Metric Viewer~\citep{steele2018vis}, compare-mt~\citep{neubig2019compare}, etc. Quite a few of them are collections of command-line scripts for metric calculation, which lack visualization to better present and interpret the scores. Some of them are able to generate static HTML reports to present charts and examples, but they do not allow updating visualization options interactively. MT-ComparEval is the only software we found that has an interactive user interface. It is, however, written in PHP, which unlike Python lacks a complete NLP eco-system. The number of metrics it supports is also limited and the software is no longer being actively developed. Support of multiple references is not a prevalent standard across all the softwares we investigated, and none of them supports multiple sources or sources in non-text modalities such as image, audio and video. Almost all the metric implementations are single-processed, which cannot leverage the multiple cores in modern CPUs for speedup and better scalability.

With the above limitations identified, we want to provide a unified and scalable solution that gets rid of all those constraints and is enhanced with a user-friendly interface as well as the latest NLP technologies.
In this paper, we present VizSeq, a visual analysis toolkit for a wide variety of text generation tasks, which can be used for: 1) instance-level and corpus-level system error analysis; 2) exploratory dataset analysis; 3) public data hosting and system benchmarking. It provides visualization in Jupyter notebook or a web app interface. A system overview can be found in Figure~\ref{fig:vizseq_overview}. We open source the software 
at \url{https://github.com/facebookresearch/vizseq}.
\begin{table}[t]
    \centering
    \small
    \begin{tabularx}{\columnwidth}{lX}
    \toprule
    Source Type & Example Tasks \\
    \midrule
    Text & machine translation, text summarization, dialog generation, grammatical error correction, open-domain question answering \\
    Image & image captioning, visual question answering, optical character recognition \\
    Audio & speech recognition, speech translation \\
    Video & video description \\
    Multimodal & multimodal machine translation \\
    \bottomrule
    \end{tabularx}
    \caption{Example text generation tasks supported by VizSeq. The sources can be from various modalities.}
    \vspace{-10pt}
    \label{tab:text_gen_tasks}
\end{table}

\begin{table}[t]
    \centering
    \small
    \begin{tabularx}{\columnwidth}{lXXXX}
    \toprule
    Metrics & VizSeq & compare-mt & nlg-eval & MT-Compar-Eval \\
    \midrule
    BLEU & \checkmark & \checkmark  & \checkmark & \checkmark \\
    chrF & \checkmark & \checkmark  & & \\
    METEOR & \checkmark & \checkmark  & \checkmark & \\
    TER & \checkmark & & & \checkmark \\
    RIBES & \checkmark & \checkmark & & \\
    GLEU & \checkmark & & & \\
    NIST & \checkmark &  & & \\
    ROUGE & \checkmark & \checkmark  & \checkmark & \\
    CIDEr & \checkmark &  & \checkmark & \\
    WER & \checkmark & \checkmark  & & \\
    \midrule
    LASER & \checkmark &  &   & \\
    BERTScore & \checkmark &  & & \\
    \bottomrule
    \end{tabularx}
    \caption{Comparison of VizSeq and its counterparts on n-gram-based and embedding-based metric coverage.}
    \label{tab:metrics_list}
    \vspace{-10pt}
\end{table}

\section{Main Features of VizSeq}
\subsection{Multimodal Data and Task Coverage}
VizSeq has built-in support for multiple sources and references. The number of references is allowed to vary across different examples, and the sources are allowed to come from different modalities, including text, image, audio and video. This flexibility enables VizSeq to cover a wide range of text generation tasks and datasets, far beyond the scope of machine translation, which previous softwares mainly focus on. Table \ref{tab:text_gen_tasks} provides a list of example tasks supported by Vizseq.
\begin{figure}[t]
    \centering
    \small
    \includegraphics[width=0.48\textwidth]{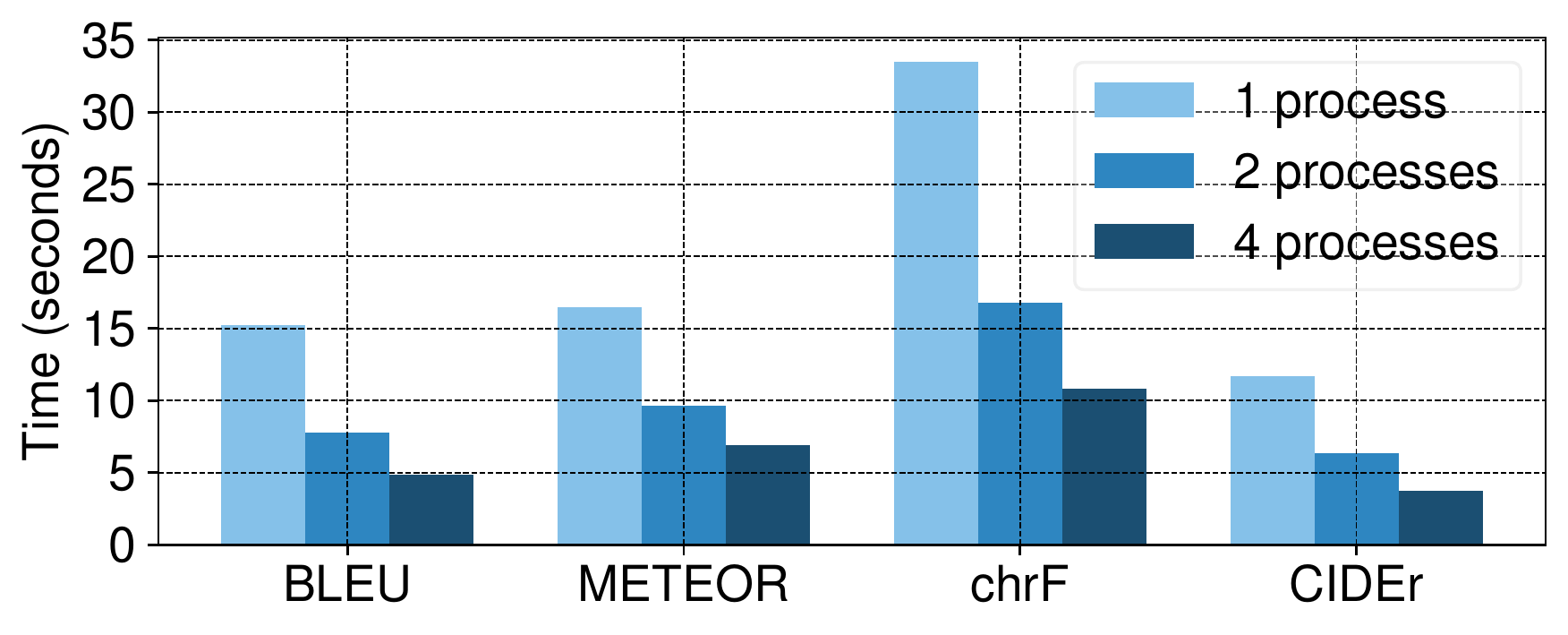}
    \caption{VizSeq implements metrics with multiprocessing speedup. Speed test is based on a 36k evaluation set for BLEU, METEOR and chrF, and a 41k one for CIDEr. CPU: Intel Core i7-7920HQ @ 3.10GHz}
    \label{fig:scorer_speedup}
    \vspace{-10pt}
\end{figure}

\subsection{Metric Coverage and Scalability}
Table~\ref{tab:metrics_list} shows the comparison of VizSeq and its counterparts on metric coverage.

\paragraph{N-gram-based metrics} To the extent of our knowledge, VizSeq has the best coverage of common n-gram-based metrics, including BLEU~\citep{papineni2002bleu}, NIST~\citep{doddington2002automatic}, METEOR~\citep{banerjee2005meteor}, TER~\citep{snover2006study}, RIBES~\citep{isozaki2010automatic}, chrF~\citep{popovic2015chrf} and GLEU~\citep{wu2016google} for machine translation; ROUGE~\citep{lin2004rouge} for summarization and video description; CIDEr~\citep{vedantam2015cider} for image captioning; and word error rate for speech recognition. 

\paragraph{Embedding-based metrics} N-gram-based metrics have difficulties in capturing semantic similarities since they are usually based on exact word matches. As a complement, VizSeq also integrates latest embedding-based metrics such as BERTScore~\citep{zhang2019bertscore} and LASER~\citep{artetxe2018massively}. This is rarely seen in the counterparts.

\paragraph{Scalability} We re-implemented all the n-gram-based metrics with multiprocessing, allowing users to fully utilize the power of modern multi-core CPUs. We tested our multi-process versions on large evaluation sets and observed significant speedup against original single-process ones (see Figure~\ref{fig:scorer_speedup}). VizSeq's embedding-based metrics are implemented using PyTorch~\citep{paszke2017automatic} framework and their computation is automatically parallelized on CPU or GPU by the framework.

\paragraph{Versatility} VizSeq's rich metric collection is not only available in Jupyter notebook or in the web app, it can also be used in any Python scripts. A typical use case is periodic metric calculation during model training. VizSeq's implementations save time, especially when evaluation sets are large or evaluation is frequent. To allow user-defined metrics, we designed an open metric API, whose definition can be found in Figure~\ref{fig:vizseq_scorer_api}. 
\begin{figure}[t]
    \centering
    \small
    \begin{lstlisting}[language=Python]
from vizseq.scorers import register_scorer

@register_scorer('metric name')
def calculate_score(
    hypothesis: List[str],
    references: List[List[str]],
    n_processes: int = 2,
    verbose: bool = False
) -> Tuple[float, List[float]]:
    return corpus_score, sentence_scores\end{lstlisting}
    \caption{VizSeq metric API. Users can define and register their new metric by implementing this function.}
    \label{fig:vizseq_scorer_api}
    \vspace{-15pt}
\end{figure}
\begin{figure}[t]
    \centering
    \small
    \includegraphics[width=0.48\textwidth]{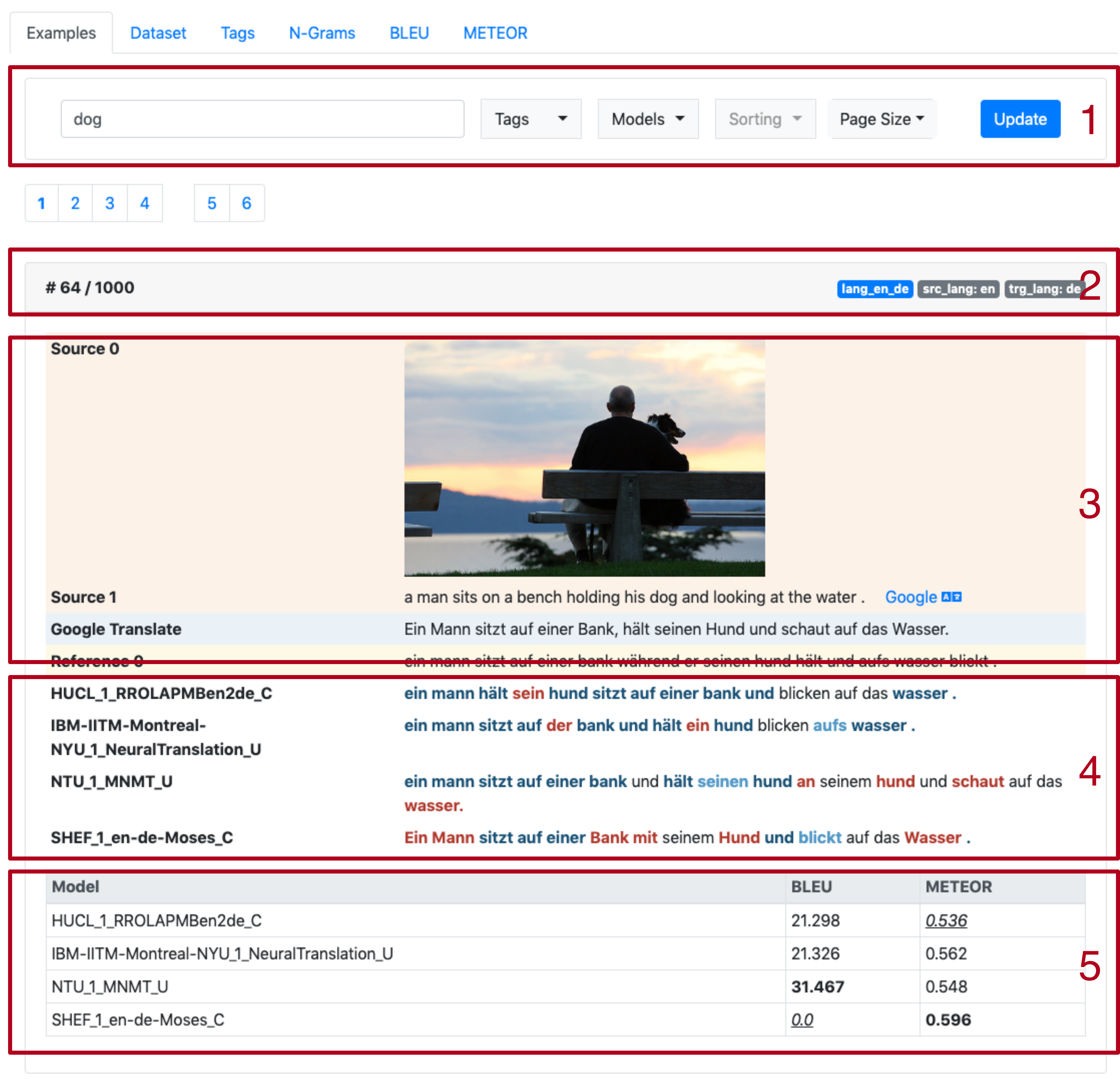}
    \caption{VizSeq example viewing. (1) keyword search box, tag and model filters, sorting and page size options; (2) left: example index, right: user-defined tags (blue) and machine-generated tags (grey); (3) multimodal sources and Google Translate integration; (4) model predictions with highlighted matched (blue) and unmatched (red) n-grams; (5) sentence-level scores (highest ones among models in boldface, lowest ones in italics with underscore). }
    \label{fig:vizseq_example_tab}
    \vspace{-4mm}
\end{figure}

\begin{figure}[t]
    \centering
    \small
    \includegraphics[width=0.48\textwidth]{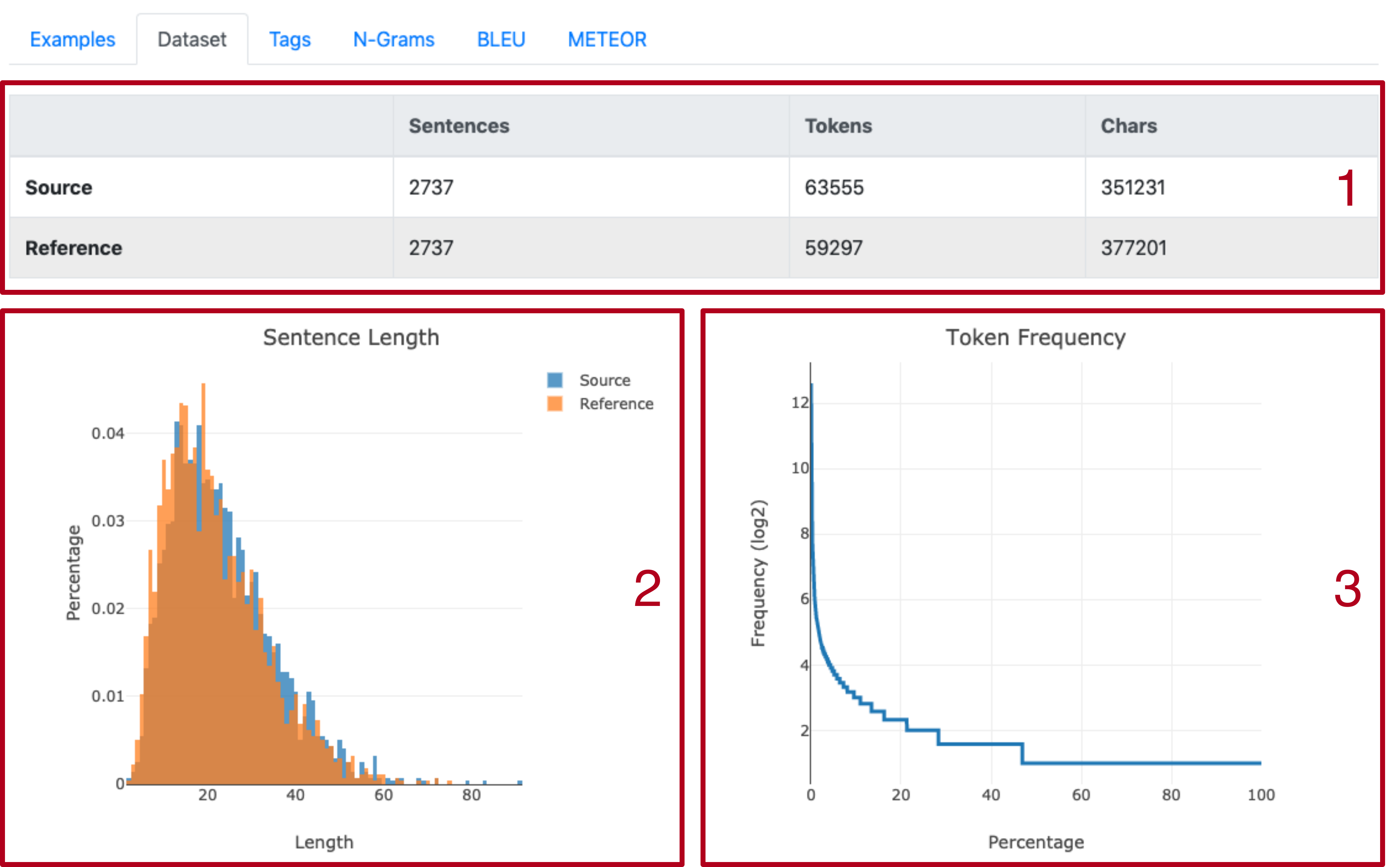}
    \caption{VizSeq dataset statistics. (1) sentence, token and character counts for source and reference sentences; (2) length distributions of source and reference sentences; (3) token frequency distribution. Plots are zoomable and exportable to SVG or PNG images.}
    \label{fig:vizseq_dataset_tab}
\end{figure}

\begin{figure}[t]
    \centering
    \small
    \includegraphics[width=0.48\textwidth]{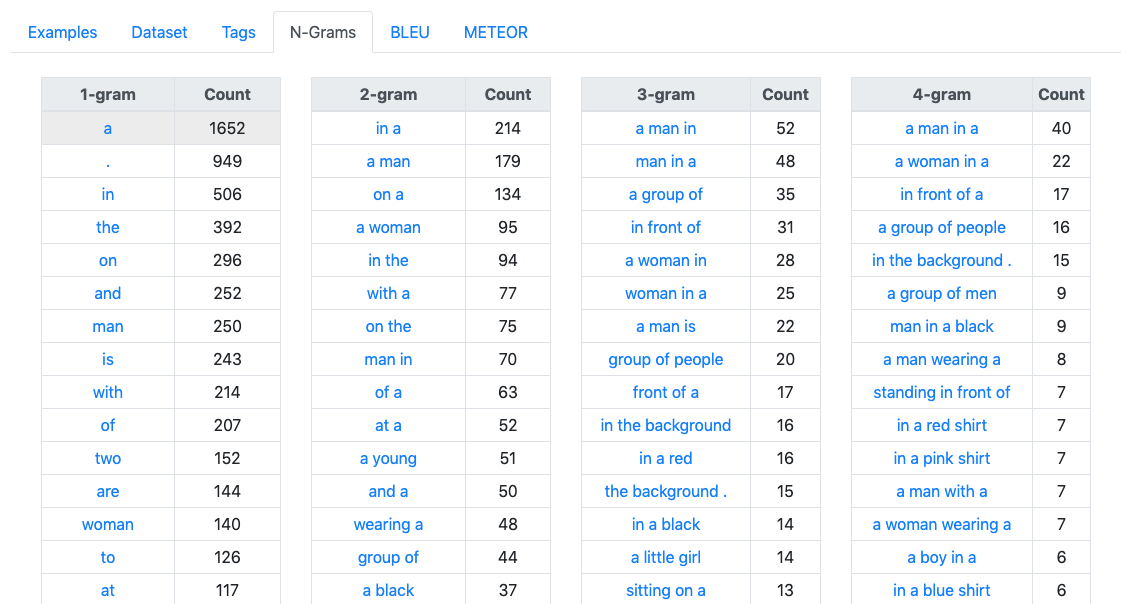}
    \caption{VizSeq dataset statistics: most frequent n-grams (n=1,2,3,4). Each listed n-gram is clickable to show associated examples in the dataset.}
    \label{fig:vizseq_ngram_tab}
    \vspace{-4mm}
\end{figure}

\begin{figure}[t]
    \centering
    \small
    \includegraphics[width=0.48\textwidth]{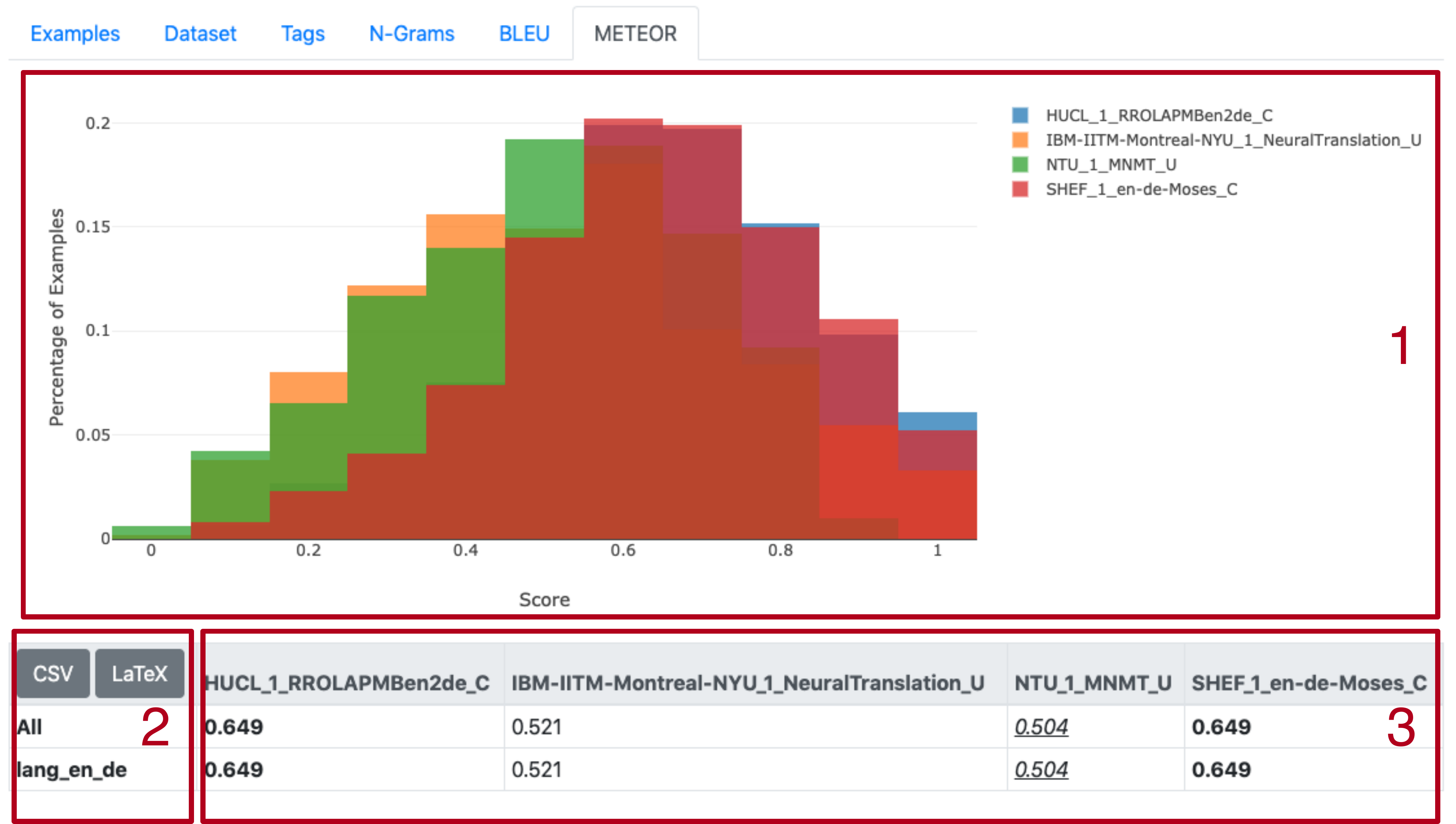}
    \caption{VizSeq corpus-level metric viewing. (1) distributions of sentence-level scores by models; (2) one-click export of tabular data to CSV and \LaTeX{} (copied to clipboard); (3) corpus-level and group-level (by sentence tags) scores (highest ones among models in boldface, lowest ones in italics with underscore).}
    \label{fig:vizseq_meteor_tab}
\end{figure}

\begin{figure}[t]
    \centering
    \small
    \includegraphics[width=0.48\textwidth]{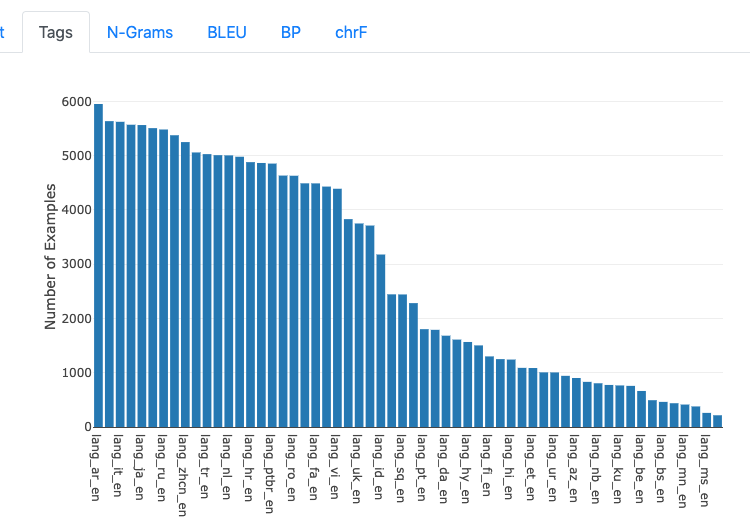}
    \caption{VizSeq sentence tag distribution view. In this example, tags are source-target language directions in a multilingual machine translation dataset.}
    \label{fig:vizseq_tag_dist}
    \vspace{-4mm}
\end{figure}

\begin{figure}[t]
    \centering
    \small
    \includegraphics[width=0.48\textwidth]{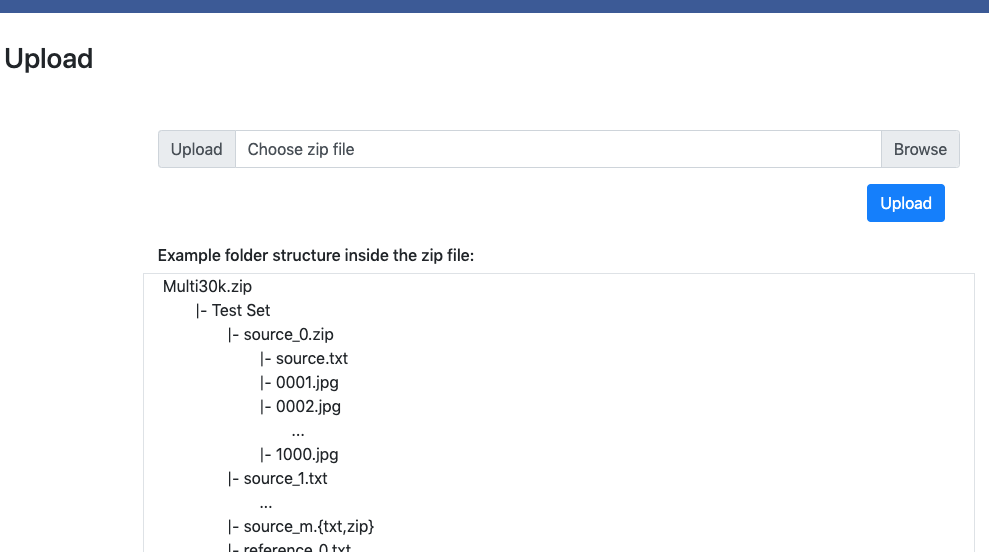}
    \caption{VizSeq data uploading. Users need to organize the files by given folder structures and pack them into a zip file for upload. VizSeq will unpack the files to the data root folder and perform integrity checks.}
    \label{fig:vizseq_screenshot_upload}
\end{figure}

\begin{figure}[t]
    \centering
    \small
    \includegraphics[width=0.48\textwidth]{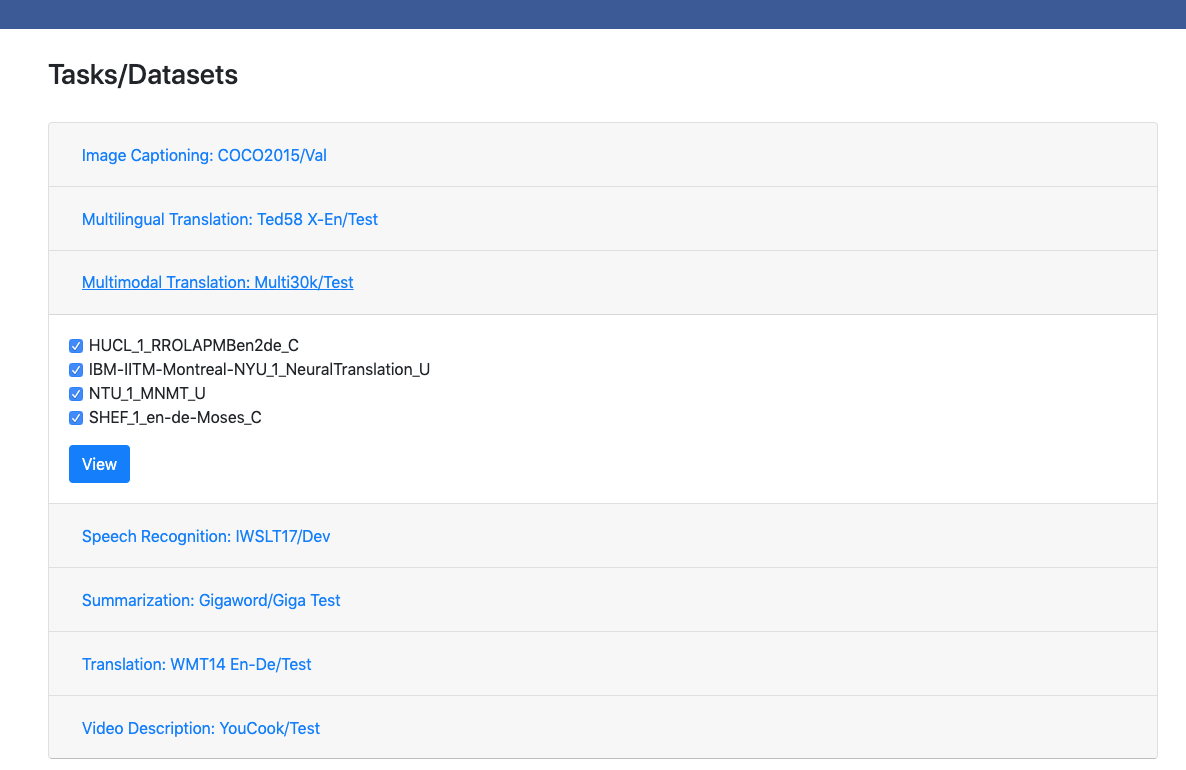}
    \caption{VizSeq task/dataset browsing. Users need to select a dataset and models of interest to proceed to the analysis module.}
    \label{fig:vizseq_screenshot_task}
    \vspace{-15pt}
\end{figure}

\subsection{User-Friendly Interface}
Given the drawbacks of simple command-line interface and static HTML interface, we aim at visualized and interactive interfaces for better user experience and productivity. VizSeq provides visualization in two types of interfaces: Jupyter notebook and web app. They share the same visual analysis module (Figure~\ref{fig:vizseq_example_tab}). The web app interface additionally has a data uploading module (Figure~\ref{fig:vizseq_screenshot_upload}) and a task/dataset browsing module (Figure~\ref{fig:vizseq_screenshot_task}), while the Jupyter notebook interface gets data directly from Python variables. The analysis module includes the following parts.

\paragraph{Example grouping} VizSeq uses sentence tags to manage example groups (data subsets of different interest, can be overlapping). It contains both user-defined and machine-generated tags (e.g. labels for identified languages, long sentences, sentences with rare words or code-switching). Metrics will be calculated and visualized by different example groups as a complement to scores over the entire dataset.

\paragraph{Example viewing} VizSeq presents examples with various sentence-level scores and visualized alignments of matched/unmatched reference n-grams in model predictions. It also has Google Translate integration to assist understanding of text sources in unfamiliar languages as well as providing a baseline translation model. Examples are listed in multiple pages (bookmarkable in web app) and can be sorted by various orders, for example, by a certain metric or source sentence lengths. Tags or n-gram keywords can be used to filter out examples of interest.

\paragraph{Dataset statistics} VizSeq provides various corpus-level statistics, including: 1) counts of sentences, tokens and characters; 2) source and reference length distributions; 3) token frequency distribution; 4) list of most frequent n-grams (with links to associated examples); 5) distributions of sentence-level scores by models (Figure~\ref{fig:vizseq_dataset_tab},~\ref{fig:vizseq_ngram_tab} and~\ref{fig:vizseq_meteor_tab}). Statistics are visualized in zoomable charts with hover text hints.

\paragraph{Data export} Statistics in VizSeq are one-click exportable: charts into PNG or SVG images (with users' zooming applied) and tables into CSV or \LaTeX{} (copied to clipboard). 

\subsection{Data Management and Public Hosting}
VizSeq web app interface gets new data from the data uploading module (Figure~\ref{fig:vizseq_screenshot_upload}) or a RESTful API.
Besides local deployment, the web app back-end can also be deployed onto public servers and provide a general solution for hosting public benchmarks on a wide variety of text generation tasks and datasets.

In VizSeq, data is organized by special folder structures as follows, which is easy to maintain:

{\small
\begin{lstlisting}[numbers=none]
<task>/<eval set>/source_*.{txt,zip}
<task>/<eval set>/reference_*.txt
<task>/<eval set>/tag_*.txt
<task>/<eval set>/<model>/prediction.txt
<task>/<eval set>/__cfg__.json
\end{lstlisting}
}

When new data comes in, scores, n-grams and machine-generated tags will be pre-computed and cached onto disk automatically. A file monitoring and versioning system (based on file hashes, sizes or modification timestamps) is employed to detect file changes and trigger necessary updates on pre-computed results. This is important for supporting evaluation during model training where model predictions change over training epochs.

\section{Example Use Cases of VizSeq}
We validate the usability of VizSeq with multiple tasks and datasets, which are included as examples in our Github repository:
\begin{itemize}[leftmargin=*]
    \setlength\itemsep{0.16em}
    \item WMT14 English-German\footnote{\url{http://www.statmt.org/wmt14/translation-task.html}}: a classic medium-size dataset for bilingual machine translation.
    \item Gigaword\footnote{\url{https://github.com/harvardnlp/sent-summary}}: a text summarization dataset.
    \item COCO captioning 2015~\citep{lin2014microsoft}: a classic image captioning dataset where VizSeq can present source images with text targets.
    \item WMT16 multimodal machine translation task 1\footnote{\url{https://www.statmt.org/wmt16/multimodal-task.html}}: English-German translation with an image the sentences describe. VizSeq can present both text and image sources, and calculate the official BLEU, METEOR and TER metrics.
    \item Multilingual machine translation on TED talks dataset~\citep{Ye2018WordEmbeddings}: translation from 58 languages into English. VizSeq can use language directions as sentence tags to generate score breakdown by languages. The test set has as many as 165k examples, where VizSeq multi-process scorers run significantly faster than single-process ones. The integrated Google Translate can help with understanding source sentences in unfamiliar languages. 
    \item IWSLT17 English-German speech translation\footnote{\url{https://sites.google.com/site/iwsltevaluation2017}}: VizSeq can present English audios with English transcripts and German text translations.
    \item YouCook~\citep{das2013thousand} video description: VizSeq enables inspecting generated text descriptions with presence of video contents.
\end{itemize}

\section{Related Work}
With the thrive of deep learning, task-agnostic visualization toolkits such as Tensorboard\footnote{\url{https://github.com/tensorflow/tensorboard}}, visdom\footnote{\url{https://github.com/facebookresearch/visdom}} and TensorWatch\footnote{\url{https://github.com/microsoft/tensorwatch}}, have emerged in need of monitoring model statistics and debugging model training. Model interpretability is another motivation for visualization. In NLP, softwares have been developed to interpret model parameters (e.g. attention weights) and inspect prediction generation process: LM~\citep{rong2016visual}, OpenNMT visualization tool~\citep{klein2018opennmt} and Seq2Seq~\citep{strobelt2019s}. For machine translation, toolkits are made to perform metric calculation and error analysis: ibleu~\citep{madnani2011ibleu}, MTEval~\footnote{\url{https://github.com/odashi/mteval}}, MT-ComparEval~\citep{klejch2015mt}, nlg-eval~\citep{sharma2017nlgeval}, Vis-Eval Metric Viewer~\citep{steele2018vis} and compare-mt~\citep{neubig2019compare}.

\section{Conclusion}
In this paper, we present VizSeq, a visual analysis toolkit for \{text, image, audio, video\}-to-text generation system evaluation, dataset analysis and benchmark hosting. It is accessible as a web app or a Python package in Jupyter notebook or Python scripts. VizSeq is currently under active development and our future work includes: 1) enabling image-to-text and video-to-text alignments; 2) adding human assessment modules; 3) integration with popular text generation frameworks such as fairseq\footnote{\url{https://github.com/pytorch/fairseq}}, opennmt\footnote{\url{https://github.com/OpenNMT/OpenNMT-py}} and tensor2tensor\footnote{\url{https://github.com/tensorflow/tensor2tensor}}.

\section*{Acknowledgments}
We thank the anonymous reviewers for their comments. We also thank Ann Lee and Pratik Ringshia for helpful discussions on this project.

\bibliography{emnlp-ijcnlp-2019}
\bibliographystyle{acl_natbib}

\appendix

\end{document}